\documentclass[letterpaper, 12pt]{l4dc2025}

\title[Learning Feasible Transitions for Efficient Contact Planning]{Learning Feasible Transitions for Efficient Contact Planning}

\usepackage{graphicx}
\usepackage{times}
\usepackage{stmaryrd}
\usepackage{booktabs}
\usepackage{algorithm}
\usepackage{algpseudocode}
\usepackage{graphicx} 
\usepackage{multirow}

\usepackage{xcolor}

\SetCommentSty{mycommfont}

\newcommand{\stateMCTS} {\mathbf{s}}
\newcommand{\stateR} {\mathbf{S}}
\newcommand{\heuristic}{h}

\newcommand{\action}{\mathbf{a}}

\newcommand{\outputs}{\mathbf{y}}
\newcommand{\stones}{\mathcal{C}}
\newcommand{\eeffpos}{\mathcal{E}}
\newcommand{\classifier}{\mathtt{c}}
\newcommand{\statepred}{\mathtt{m}}
\newcommand{\targetnet}{\mathtt{t}}

\SetKw{Continue}{continue}
\SetKw{Break}{break}

\usepackage{amsmath}

\author{%
 \Name{Rikhat Akizhanov} \Email{rikhat.akizhanov@mbzuai.ac.ae}\\
 \addr Mohamed bin Zayed University of Artificial Intelligence (MBZUAI), UAE
 \AND
 \Name{Victor Dhédin} \Email{victor.dhedin@tum.de}\\
 \addr Munich Institute of Robotics and Machine Intelligence, Technical
University of Munich, Germany
 \AND
 \Name{Majid Khadiv} \Email{majid.khadiv@tum.de}\\
 \addr Munich Institute of Robotics and Machine Intelligence, Technical
University of Munich, Germany
 \AND
 \Name{Ivan Laptev} \Email{ivan.laptev@mbzuai.ac.ae}\\
 \addr Mohamed bin Zayed University of Artificial Intelligence (MBZUAI), UAE
}

\begin{document}

\maketitle

\begin{abstract}%
In this paper, we propose an efficient contact planner for quadrupedal robots to navigate in extremely constrained environments such as stepping stones. The main difficulty in this setting stems from the mixed nature of the problem, namely discrete search over the steppable patches and continuous trajectory optimization. To speed up the discrete search, we study the properties of the transitions from one contact mode to another. In particular, we propose to learn a dynamic feasibility classifier and a target adjustment network. The former predicts if a contact transition between two contact modes is dynamically feasible. The latter is trained to compensate for misalignment in reaching a desired set of contact locations, due to imperfections of the low-level control. We integrate these learned networks in a Monte Carlo Tree Search (MCTS) contact planner. Our simulation results demonstrate that training these networks with offline data significantly speeds up the online search process and improves its accuracy.%
\end{abstract}

\begin{keywords}%
  legged robot, contact planning, motion planning, deep learning%
\end{keywords}

\section{Introduction}
The main advantage of legged robots over wheeled robots is their ability to traverse highly irregular and sparse terrains. However, generating motions on such terrains is particularly challenging for both model-based and learning-based methods.
Model-based method has two major approaches which either simply consider the geometry of the environment as a nonlinear constraint in the optimal control problem \cite{tassa2012synthesis,mordatch2012discovery,posa2014direct,winkler2018gait}, or segment the environment to a set of disconnected convex patches \cite{deits2015computing} that adds a discrete variable of selecting the proper patch into the motion generation problem, resulting in a mixed-integer optimization problem \cite{deits2014footstep, aceituno2017simultaneous,lin2020robusthumanoidcontactplanning,ponton2021efficient}.
Both approaches are limited in the case of locomotion in highly constrained environments like stepping stones. The first approach is often stuck in local minima for terrains with extreme height variations \cite{winkler2018gait}. The second approach often relies on simplifications that do not take into account the full dynamics of the robot \cite{deits2014footstep, ponton2021efficient,tonneau2020slim}, which limits the robot to perform quasi-static maneuvers.


Environments with sparse steppable locations also pose problems for learning-based methods~\cite{grandia2023perceptive} and reinforcement learning (RL) methods in particular. 
To address the challenge of sparse environments, recent methods deploy teacher/student frameworks~\cite{zhang2023learning} and curriculum learning where the difficulty of the problem is gradually increased over several training rounds.
As an alternative to pure learning approaches, other recent methods deploy a hybrid strategy combining model-based and learning-based methods.
For example, \cite{jenelten2024dtc} deploys trajectory optimization to guide RL exploration by providing dense reward signals as in~\cite{bogdanovic2022model}. However, this approach is expensive as the decision-making problem must be solved twice (once using TO and once RL) and still requires numerous samples.
Several other methods attempt to speed up the planning at runtime by learning optimal control outputs during training.
%
In \cite{lin2020robusthumanoidcontactplanning,meduri2021deepq}, a network is trained to predict robot capturability under disturbances and then used at runtime to produce footstep sequences.
In \cite{Bratta2024}, a model is trained to rank promising footstep locations based on a given cost function. At runtime, the model is used jointly with trajectory optimization as an acyclic online planner to select the best foothold location.
Although these two methods enable successful navigation in sparse environments, they provide quasi-static solutions and do not account for imperfections of the low-level control as well as full dynamics of the robot, which hinders performance in extremely constrained environments.


In this work, we address the challenge of contact planning for dynamic maneuvers of legged robots in highly constrained environments
by learning both dynamic transitions and low-level control compensation.
Our approach makes use of Monte Carlo Tree Search (MCTS), a method recently demonstrated to scale effectively and outperform Mixed-Integer Quadratic Programming (MIQP) in dexterous manipulation \cite{zhu2023efficient} and gait discovery in locomotion \cite{amatucci2022monte}. 
Our work extends the framework of \cite{dhédin2024diffusionbasedlearningcontactplans} where MCTS is used jointly with whole-body nonlinear model predictive control (NMPC) to find dynamically feasible contact plans on stepping stones. 
In particular, we propose to learn a dynamic feasibility classifier to predict if a contact transition between two contact modes is dynamically feasible.
We also introduce a target adjustment network
to compensate for imperfections of the low-level control. 
We experimentally compare our method to \cite{dhédin2024diffusionbasedlearningcontactplans} and demonstrate significantly increased success rate and speed of our approach when tested in various randomized stepping stones environments on a Go2 quadruped platform.
We also compare our framework with the state-of-the-art RL locomotion policy based on \cite{zhang2023learning} and show that our method enables locomotion on very challenging terrain where the RL policy fails.
In summary, this work proposes the following contributions:
\begin{itemize}
     \item 
    We propose a dynamic state predictor that estimates the state of the robot for a given contact mode transition, and a feasibility classifier to evaluate if this transition is dynamically feasible. This enables efficient pruning in the tree search. 
    
\vspace{-.1cm}
    \item We introduce a learning-based method to compensate for inaccuracies of a given low-level controller by regressing its errors with respect to target contact locations.

\vspace{-.1cm}
    \item We propose to improve the search efficiency of MCTS by incorporating \textit{safety} and \textit{accuracy} heuristics, based on the confidence level of the learned classifier and residual networks.  

\vspace{-.1cm}
    \item Our extensive experiments in randomized environments demonstrate advantages of the proposed approach in terms of speed and accuracy compared to the state of the art. 
    
\end{itemize}

The rest of the paper is organized as follows. We first formulate the problem of navigation on stepping stones and present the general MCTS framework in Section \ref{sec:formulation}. We then describe the proposed improvements in Section \ref{sec:method}. We evaluate the performance of our algorithm and compare it to other baselines in Section \ref{sec:results}. Section \ref{sec:conclusions} summarizes our findings and outlines future work.

\section{Problem formulation}
\label{sec:formulation}

\begin{figure}
    \centering
    \includegraphics[width=1.0\linewidth]{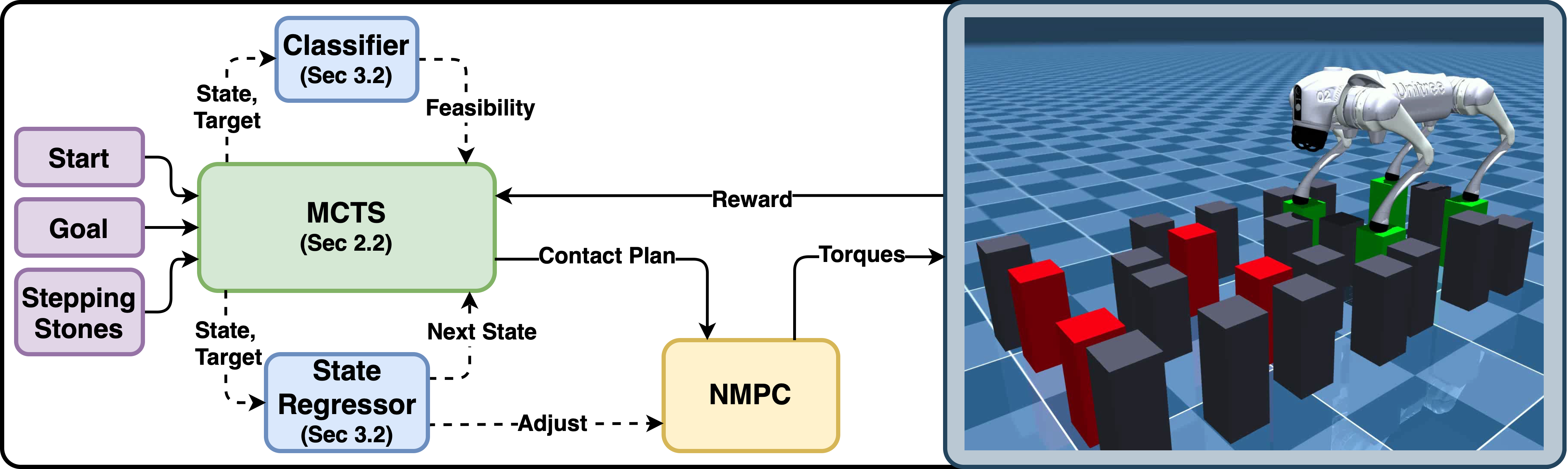}
    \vspace{-5mm}
    \caption{Framework overview with the block diagram. In the MCTS (Section \ref{sec:contact_planning_MCTS}), a state predictor and a feasibility classifier are jointly called to prune for paths that are likely not to be dynamically feasible (Section \ref{sec:classifier}). While running the NMPC in simulation to check for the dynamic feasibility of a contact plan, target contact locations given to the NMPC are adjusted to compensate for the low-level controller inaccuracies (Section \ref{sec:offsets}).}
    \label{fig:env-example}
    \vspace{-5mm}
\end{figure}

We address the problem of dynamic locomotion on stepping stones with a legged robot.
Our problem consists of reaching a goal position from a start position in a given stepping stones environment (Fig. \ref{fig:env-example}). The 
output is a sequence of stepping stones to step on in order to reach the goal, as well as low-level adjustments to land robot closer to stones' center. 

\subsection{Notations}

We consider a set of contact locations $\stones^W$ in world frame $W$ that are the centers of $N_{stones}$ stepping stones of radius $r_{stones}$. A state in the tree search $\stateMCTS$ corresponds to a set of stepping stones for each $N_e$ end-effectors, denoted by discrete indices ($\stateMCTS \in \llbracket 1, N_{stones} \rrbracket ^{N_e}$). The goal state is noted $\stateMCTS^g$. $\stones^W_{\stateMCTS} \in \mathbb{R}^{{N_e} \times 3}$ corresponds to the $N_e$ contact locations in $W$ associated to the state $\stateMCTS$. $\stones^W_{\stateMCTS}[j] \in \mathbb{R}^{3}$ correspond to the location of the $j$th end effector.

We denote $\stateR = [\mathbf{p}^{W}_{B}, \mathbf{q}^{W}_{B}, \mathbf{q}_j, \mathbf{v}^{W}_{B}, \mathbf{w}^{W}_{B}]$ the robot state. With $B$ the base frame, $\mathbf{p}^{W}_{B}$ and $\mathbf{q}^{W}_{B}$ the position and orientation (as a quaternion) of the base, $\mathbf{q}_j$ the joint position, $\mathbf{v}^{W}_{B}$ the linear velocity and $\mathbf{w}^{W}_{B}$ the angular velocity. In the following, $\Bar{\stateR}$ is the state $\stateR$ without the absolute position. $\eeffpos^{B} \in \mathbb{R}^{{N_e} \times 3}$ corresponds to the end-effectors locations in the base frame.


\subsection{Contact planning with MCTS}
\label{sec:contact_planning_MCTS}

We formalize our contact planning problem as a Markov Decision Process (MDP). In this MDP, a state $\stateMCTS$ is a set of contact locations for the end-effectors, represented by their corresponding discrete stepping stones index. The action $\action$ selects the next locations for each end-effector and brings the system to a new state $\stateMCTS ' = f(\stateMCTS,\action)$.
To solve this problem, we follow the MCTS formulation in \cite{dhédin2024diffusionbasedlearningcontactplans} and create a search tree $\mathcal{T} = (\mathcal{V}. \eeffpos)$, where the set of nodes $\mathcal{V}$ contains the visited states and the set of edges contains the visited transitions ${(\stateMCTS \overset{\action}{\rightarrow}\stateMCTS ')}$. Each transition maintains the state-action value $Q(\stateMCTS, \action)$ and the number of visits $N(\stateMCTS, \action)$. 

MCTS grows a search tree iteratively using the following steps: the \textbf{Selection} phase begins at the root node (initial state) and successively chooses child nodes until a leaf node is reached (an unexpanded node or terminal state). If all children of a node are expanded, a child node based on the highest Upper Confidence Bound (UCB) score is selected \cite{kocsis2006bandit}. The \textbf{Expansion} phase adds the successor states to the tree by enumerating all possible actions, if the selected state is not terminal. The \textbf{Simulation} phase performs random actions from one of the successor states for a predefined number of steps and evaluates the reward at the end of this phase. The \textbf{Back-propagation} phase updates the state-action values and visit counts for all states along the selected and expanded path, using the reward obtained from the simulation.

Once a sequence from start to goal is found, NMPC (we use the formulation in \cite{meduri2023biconmp}) is rolled-out in simulation to evaluate the dynamic feasibility of the whole sequence. For each transition ${(\stateMCTS \overset{\action}{\rightarrow}\stateMCTS ')}$ in the sequence, the contact locations for each end-effector corresponding to state $\stateMCTS'$ are set as the new target contacts in the NMPC. If the whole plan is not feasible (the robot didn't reach the goal), the opposite of the reward is back-propagated. This penalizes transitions leading to failure, as it lowers the value and the UCB score of the nodes in the infeasible sequence.
The NMPC is not used to check the dynamic feasibility of each transition in the tree expansion, as it would be too time-consuming. It is only used to check a full path to the terminal state, if such a kinematically feasible path is found.

To speed up the search, \cite{dhédin2024diffusionbasedlearningcontactplans} proposed a few adjustments.
In the selection and simulation step, a heuristic guides the search towards states closer to the goal (shortest path).
In the expansion step, only a kinematically feasible set of contact locations is expanded to prune the tree: states corresponding to a configuration where the legs are crossed or too far from the current contact locations are pruned. We refer to this as \textit{kinematic} pruning.
It will be our baseline in the experiments and corresponds to the algorithm proposed by \cite{dhédin2024diffusionbasedlearningcontactplans}.

\section{Method}
\label{sec:method}

In this section, we propose three major modifications to the MCTS contact planner to improve its accuracy and to speed up the search process, see Fig.~\ref{fig:env-example} for an overview.

As the first modification, we propose to train a classifier $\classifier$ and a state predictor $\statepred$ that are evaluated during the search in the expansion step to prune the dynamically infeasible transitions. The classifier takes as input the robot state $\Bar{\stateR}$ and the contact locations of the transition, i.e. $\stones^B_{\stateMCTS}$ and $\stones^B_{\stateMCTS'}$. During the search, the state of the robot after a transition is unknown. Therefore, $\Bar{\stateR}$ is predicted jointly by  a state predictor $\statepred$.
Only states with a classification score higher than a threshold $t_{feasible}$ are considered feasible. This is what we call \textit{dynamic} pruning in the rest of the paper.
Details about the classifier and predictor networks are given in Sections \ref{sec:classifier}.

As the second modification, we introduce a \textit{target adjustment network} $\targetnet$ to reduce the inaccuracies of the NMPC controller during the simulation. The network is trained so that the achieved contact locations (see Fig. \ref{fig:offset-illustration}) are closer to the target locations given to the NMPC, as presented in Section \ref{sec:offsets}. This is notably useful in the case of small stepping stones, where accuracy is needed, as demonstrated in Section \ref{sec:results:adjustment}. 

As the third modification, we propose to use the confidence of the classifier and state predictor networks to inform the search. We take into account the chances of dynamically succeeding in a transition based on the confidence level of the neural networks. We greedily use the heuristic in the selection and simulation steps of MCTS by selecting the state with maximum heuristic value. This is detailed in Section \ref{sec:heuristic}.


Algorithms \ref{alg:expansion} and \ref{alg:simulation} provide a pseudocode for the modified expansion and simulation steps. For clarity, operations on the successor states are performed in a \textit{for} loop, but in practice, they are performed in a batched manner for efficiency purposes.

To simplify the notations in the following, we note $\Bar{\stateR}_{\mathcal{P}}$ the predicted robot state associated with the search path $\mathcal{P}$. $\stones^{B_{\mathcal{P}}}_{\stateMCTS}$ corresponds to the contact locations associated with states $\stateMCTS$ in the predicted base frame $B_{\mathcal{P}}$. For a transition $\stateMCTS \rightarrow \stateMCTS'$, we note $\classifier(\stateMCTS, \stateMCTS')$ the classifier taking as inputs the predicted state $\Bar{\stateR}_{\mathcal{P}}$, the current contact locations $\stones^{B_{\mathcal{P}}}_{\stateMCTS}$ and the target contact locations $\stones^{B_{\mathcal{P}}}_{\stateMCTS'}$. We use a similar notation for $\statepred(\stateMCTS, \stateMCTS')$ and $\targetnet(\stateMCTS, \stateMCTS')$.

\RestyleAlgo{algo2e,ruled,linesnumbered,ruled,vlined,noend}
\begin{algorithm}[h!]
\caption{Pseudocode for the modified \textbf{expansion} step in MCTS}
\label{alg:expansion}
\KwIn{Tree $\mathcal{T}$, current search path $\mathcal{P}$, selected node $\stateMCTS$, state predictor network $\statepred$, feasibility classifier $\classifier$, threshold $t_{feasible}$, maximum transition length $d_{max}$}
\KwOut{Feasible successor states added to the tree}

$\stateR \gets \text{predict\_robot\_state}(\mathcal{P}, \statepred)$\;

\ForEach{$\stateMCTS' \in \text{Sucessors}(\mathcal{T}, \stateMCTS)$}{

    \tcp{kinematic pruning}\
    \uIf{$\text{crossing\_leg}(\stateMCTS') \text{ or } 
         \text{transition\_length}(\stateMCTS, \stateMCTS') > d_{max}$}{
        Discard $\stateMCTS'$; \Continue \;
     }
    
    \tcp{dynamic pruning}\
    \uIf{$\text{not dyn\_feasible}(\classifier, \stateMCTS, \stateMCTS', \stateR, t_{feasible})$}{
        Discard $\stateMCTS'$; \Continue \;
    }
    \tcp{Add $\stateMCTS'$ to the tree as a child node\;}
    $\text{add\_successor}(\mathcal{T}, \stateMCTS, \stateMCTS')$   
}
\end{algorithm}

\begin{algorithm}[h!]
\caption{Pseudocode for the modified \textbf{simulation} step in MCTS}
\label{alg:simulation}

\KwIn{Tree $\mathcal{T}$, search path $\mathcal{P}$, selected state $\stateMCTS$, heuristic function $h$, target adjustment network $\targetnet$}
\KwOut{Reward $r$}
$\mathcal{P}_{sim} \gets \mathcal{P}$

\For{$N_{sim} \text{ steps}$}{
    \tcp{Choose next state greedily and store maximum heuristic value}\
    $\stateMCTS \gets \arg\max_{\stateMCTS' \in \text{Successors}(\mathcal{T}, \stateMCTS)} h(\stateMCTS, \stateMCTS')$
    ; $h^{*} \gets \max_{\stateMCTS' \in \text{Successors}(\mathcal{T}, \stateMCTS)} h(\stateMCTS, \stateMCTS')$
    
    \tcp{Add $\stateMCTS$ to the simulation path}
    $\mathcal{P}_{sim} \gets [\mathcal{P}_{sim}, \stateMCTS]$
    
    \uIf{$\stateMCTS$ = $\stateMCTS_g$}{
        \Break
    }
}
$r \gets h^{*}$

\uIf{$\stateMCTS$ = $\stateMCTS_g$}{
    \tcp{Roll-out the NMPC in simulation over the whole contact sequence.}
    \tcp{Adjust the target contact locations at runtime with $\targetnet$}
    $sucess = \text{dynamic\_sim}(\text{NMPC}, \mathcal{P}_{sim}, \targetnet)$
    
    \uIf{$\text{not}(sucess)$}{
        $r \gets -h^{*}$
    }
}

\Return $r$\;
\end{algorithm}

\subsection{Dataset collection}
\label{sec:dataset}
To train the networks, we perform an offline data collection process with NMPC performing locomotion on flat terrain. The NMPC tries to follow a contact plan with randomized contact locations with a predefined gait. Before each gait cycle, we record:
\begin{itemize}
    \item the state of the robot base $\stateR_i$ before the start of the cycle (noted $B_i$)
    \item the current end-effector positions $\eeffpos^{B_i}_i$
    \item the next target contact locations ${\eeffpos^{*}}^{B_i}_{i+1}$
\end{itemize}

A transition is considered successful if the robot is not in collision and if the position error between the targeted and achieved contact locations is less than a threshold $e_{max}=8cm$. In that case, we assign a label $\outputs_i = 1$ to the transition, otherwise $\outputs_i = 0$.
Note that the joint velocities are not considered in the robot state $\stateR$. Considering them did not improve the performances of the classifier, as shown in Section \ref{sec:results:networks}. 


\subsection{Feasibility classifier and state predictor}
\label{sec:classifier}

The classifier $\classifier$ is trained with supervised learning on samples $([\Bar{\stateR}_i, \eeffpos^{B_i}_i, {\eeffpos^{*}}^{B_i}_{i+1}], \outputs_i)$. The network outputs the logits probability (or score) to succeed the transition.

The state predictor $\statepred$ is trained with supervised learning on samples $([\Bar{\stateR}_i, \eeffpos^{B_i}_i, \eeffpos^{B_i}_{i+1}], \Bar{\stateR}_{i+1})$, the output being here the state after transition $i$. The state predictor is only trained on transitions that succeeded ($\outputs_{i} = 1$). Note that in this case, the two set of contact locations given in input correspond to the \textit{achieved} contact locations. For the classifier, the target contact locations are given in input, as the contact state may not be achieved in cases of failure (see Fig. \ref{fig:offset-illustration}).

During the search, for a transition $\stateMCTS \rightarrow \stateMCTS'$, we provide as input to the networks $[\Bar{\stateR}_{\mathcal{P}}, \stones^{B_{\mathcal{P}}}_{\stateMCTS}, \stones^{B_\mathcal{P}}_{\stateMCTS'}]$. This requires to estimate the absolute base position $\mathbf{p}^{W}_{B_{\mathcal{P}}}$ in order to express the stepping stones locations $\stones^W$ in base frame.
However, the absolute position of the base is not predicted by the network, as all position inputs are expressed in the local frame. We propose to estimate the translation induced by transition $i$, noted $\mathbf{t}^{B_i}_{B_{i+1}}$, as one can compute the absolute base position with $\mathbf{p}^{W}_{B_{i+1}} = \mathbf{p}^{W}_{B_{i}} + \mathbf{R}^{W}_{B_{i}} \mathbf{t}^{B_i}_{B_{i+1}}$. $\mathbf{R}^{W}_{B_{i}}$ is the orientation of the base as a rotation matrix and is predicted by the network.
To do so, we first compute the end-effector positions in the predicted joint configuration with forward kinematics, noted $\Tilde{\eeffpos}^{B_i}_i$.
Then, we compute the translation that moves the base so that the end-effectors are as close as possible to the target contact locations in the current configuration i.e. the translation minimizes the average distance between $\Tilde{\eeffpos}^{B_i}_i$, and the target locations ${\eeffpos^{*}}^{B_{i}}_{i+1}$:

\vspace{-7mm}
\begin{align}\label{eq:translation}
    \mathbf{t}^{B_i}_{B_{i+1}} = \frac{1}{N_e} \sum_{j=1}^{N_e}({\eeffpos^{*}}^{B_{i}}_{i+1}[j] - \Tilde{\eeffpos}^{B_i}_i[j])
\end{align} 


\subsection{Compensating for the low-level controller inaccuracies}
\label{sec:offsets}

\begin{figure}
    \centering
    \includegraphics[width=0.5\linewidth]
    {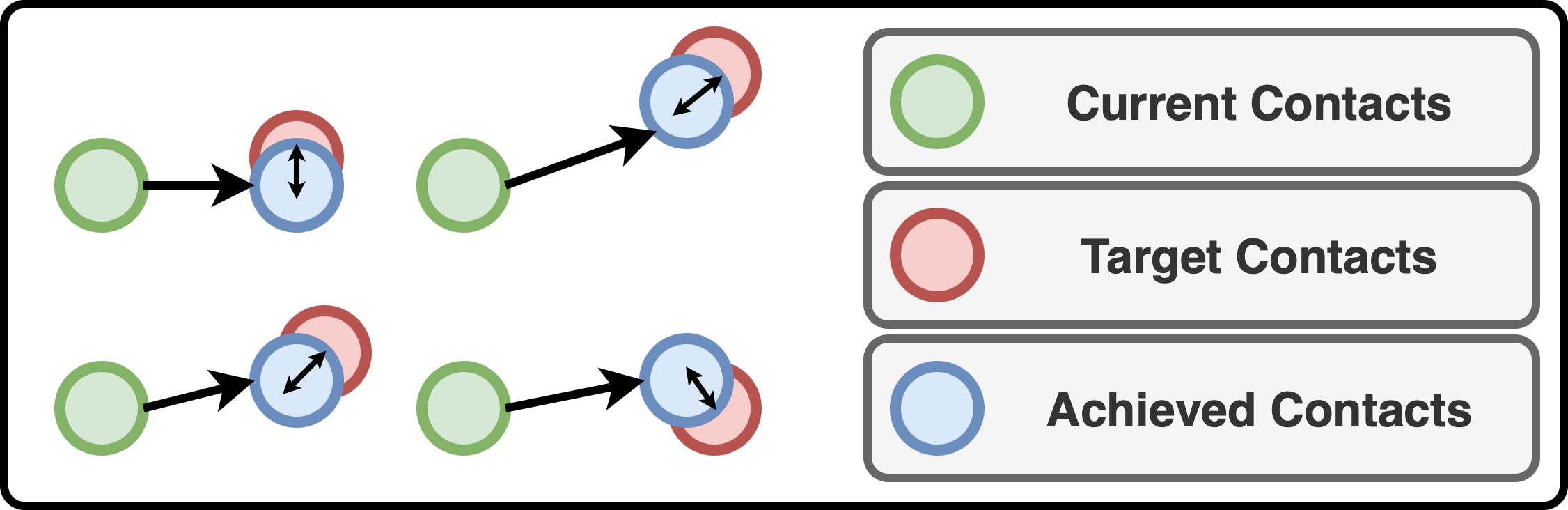}
    \vspace{-5mm}
    \caption{Illustration of current, target and achieved contacts for a jump with a quadruped robot. The NMPC doesn't precisely achieve target position. The residual is defined as the difference between target and achieved contacts.}
    \label{fig:offset-illustration}
    \vspace{-5mm}
\end{figure}

The NMPC controller is not perfectly accurate in terms of  reaching exactly the targeted contact locations. This is illustrated on Fig. \ref{fig:offset-illustration}. While this accuracy is not crucial for locomotion on flat terrain, it becomes critical to navigate on highly sparse terrains such as stepping stones. To tackle this, we trained a \textit{target adjustment network} $\targetnet$ to compensate for the inaccuracies of the low-level controller by adjusting the targeted contact positions.
The network is trained on samples $([\Bar{\stateR}_i, \eeffpos^{B_i}_i, \eeffpos^{B_i}_{i+1}], \eeffpos^{B_i}_{i+1} - {\eeffpos^{*}}^{B_i}_{i+1})$. The output of the network is the residual to add to the target ${\eeffpos^{*}}^{B_i}_{i+1}$ (input to the NMPC) in order to reach the achieved location $\eeffpos^{B_i}_{i+1}$. Indeed, if now the target is $\eeffpos^{B_i}_{i+1}$, the controller will reach exactly those locations by giving ${\eeffpos^{*}}^{B_i}_{i+1}$ as input.  

At runtime, we call the network before each gait cycle, add the residuals to the target contact locations, and provide the result to the NMPC.
Note that the state predictor network and the target adjustment network have similar inputs. In practice, we use one network that outputs both the predicted state and residuals. We refer to them as one network in the result section.

\subsection{Heuristics and reward}
\label{sec:heuristic}

The proposed heuristic is composed of three terms. The first one guides the search in the direction of the goal as introduced by \cite{dhédin2024diffusionbasedlearningcontactplans}, denoted by $\heuristic_{goal} \in [0, 1]$. The second one accounts for the chance of succeeding the desired gait, specified as $\heuristic_{safety} \in [0, 1]$. The third one accounts for the amount of correction of the low-level controller, i.e. $\heuristic_{accuracy} \in [0, 1]$. The final heuristic is defined as follows, $h = \heuristic_{goal} + \alpha \heuristic_{safety} + \beta \heuristic_{accuracy}$ with $\alpha, \beta \in \mathbb{R}$. Since $\heuristic_{goal}$ is higher for states closer to the goal, we also chose to use $h$ as our reward function. This is detailed in Algorithm~\ref{alg:simulation}. The heuristics $\heuristic_{goal}$, $\heuristic_{safety}$ and $\heuristic_{accuracy}$ are defined as follows

\vspace{-3mm}
\begin{align}\label{eq:h_goal}
    \heuristic_{goal}(\stateMCTS, \action) = \sigma_{g}(\frac{1}{N_e} \sum_{j=1}^{n_e} (1 - \frac{\vert\vert \stones^{W}_{\stateMCTS}[j] - \stones^{W}_{\stateMCTS^g}[j] \vert\vert_2}{d_{max}}) )
\end{align} 
where $d_{max}$ is the maximum distance between a pair of contact locations in the map,
\vspace{-3mm}
\begin{align}\label{eq:h_safety}
    \heuristic_{safety}(\stateMCTS, \stateMCTS') = \sigma_{s}(\classifier(\stateMCTS, \stateMCTS'))
\end{align} 
where $\classifier(\stateMCTS, \stateMCTS')$ is the classification score, as we assume the higher score implies the higher confidence of the classifier,
%
\vspace{-3mm}
\begin{align}\label{eg:h_accuracy}
    \heuristic_{accuracy}(\stateMCTS, \stateMCTS')  = \sigma_{a}(\frac{1}{\Delta_{res}(\targetnet(\stateMCTS, \stateMCTS')) + 10^{-12}})   
\end{align}
where $\Delta_{res}(\targetnet(\stateMCTS, \stateMCTS'))$ is the sum of the norm of the residual values predicted by the target adjustment network. Higher heuristics values correspond to transitions that requires less adjustment and should more likely be feasible.
$\sigma_{g}$, $\sigma_{s}$, $\sigma_{a}$ are sigmoid-based functions (noted $\sigma$) with linear scaling to map heuristics values in $[0,1]$. They are defined such as:  $\sigma_{g}(x) = \sigma(5x)$, $\sigma_{s}(x) = \sigma(x / 5)$, $\sigma_{a}(x) = \sigma(x / 5)$.

\section{Results}
\label{sec:results}



In this section we present results of the proposed framework applied to a Go2 quadruped robot. We use MuJoCo simulator~\cite{Todorov2012} and control the robot in a stepping stone environment with varying stone sizes. We provide results for two different gaits, trotting and jumping. We first detail the training procedures of the different networks in Section \ref{sec:results:networks} (the procedure is the same for both gaits). We then describe the test stepping stones environment in Section \ref{sec:results:setup}. We evaluate how dynamic pruning and correcting the inaccuracies of the low-level controller affect the performance in Sections \ref{sec:results:adjustment} and \ref{sec:results:dymamic}. We compare our method to a state-of-the-art RL approach in Section \ref{sec:results:RL}. Finally, we show in Section \ref{sec:results:safety} that the performance can be further improved using the proposed safety and accuracy
heuristics.

\subsection{Training the classifier and state predictor networks}
\label{sec:results:networks}

Both networks are Mutli-Layer Perceptron (MLP) trained on datasets recorded from the same transitions as described in Section \ref{sec:dataset}. The training and validation sets contain respectively $130000$ and $13000$ samples (different datasets for jumping and trotting with the same size). We augment the data during training by adding small Gaussian noise with a standard deviation of $\sigma_{aug}^2$ to the input, which improves the training performance. We trained the networks using the Adam optimizer with an initial learning rate of $0.001$ and an exponential learning scheduler with a rate of $0.98$. 
Training parameters, network architectures and performances are detailed in table \ref{tab:mlp_training_parameters}.

\begin{table}[ht]
\centering
\vspace{-5mm}
\caption{Training parameters and performance metrics for different MLP networks. BCE: Binary Cross-Entropy. MSE: Mean Squared Error.}
\resizebox{\textwidth}{!}{%
\begin{tabular}{@{}lcccccc|cc@{}}
\toprule
\textbf{Network}         & \textbf{N. Layers} & \textbf{H. dim.} & \textbf{Act.} & \textbf{Batch Size} & $\sigma_{aug}^2$ & \textbf{Loss} & \textbf{ROC AUC} & \textbf{Accuracy} \\ \midrule
$\mathbf{\classifier}$ (Jump) & 4                  & 64               & ReLU          & 512                 & 1e-4             & BCE           & 0.93             & 0.82             \\
$\mathbf{\classifier}$ (Trot) & 4                  & 64               & ReLU          & 512                 & 1e-4             & BCE           & 0.95             & 0.85             \\
$\statepred / \targetnet $         & 3                  & 128              & ReLU          & 512                 & 1e-4             & MSE           & -                & -                \\ \bottomrule 
\end{tabular}%
}
\vspace{-8mm}
\label{tab:mlp_training_parameters}
\end{table}

\subsection{Experimental Setup}
\label{sec:results:setup}

To evaluate the performance of our contact planner, we considered an environment with up to $N_{stones} = 35$ squared stones placed on a $7 \times 5$ grid. The stepping stones initially form a regular grid of spacing $(e_x, e_y)$ so that the feet lay on $4$ stepping stones in the initial configuration. The position of each stone is then displaced by $\epsilon_x (\frac{e_x}{2} - r) $ with $\epsilon_x \sim \mathcal{U}(-0.75, 0.75)$ in the $x$ direction (respectively for the direction $y$). $N_{removed} = 9$ stepping stones are randomly removed. Additionally, a small random noise $\epsilon_h \sim \mathcal{U}(-0.02, 0.02)$ is added to the height of the stones.

\begin{figure}[h]
    \vspace{-2mm}
    \centering
    \subfigure[Jumping Gait]{
        \includegraphics[width=0.97\textwidth]{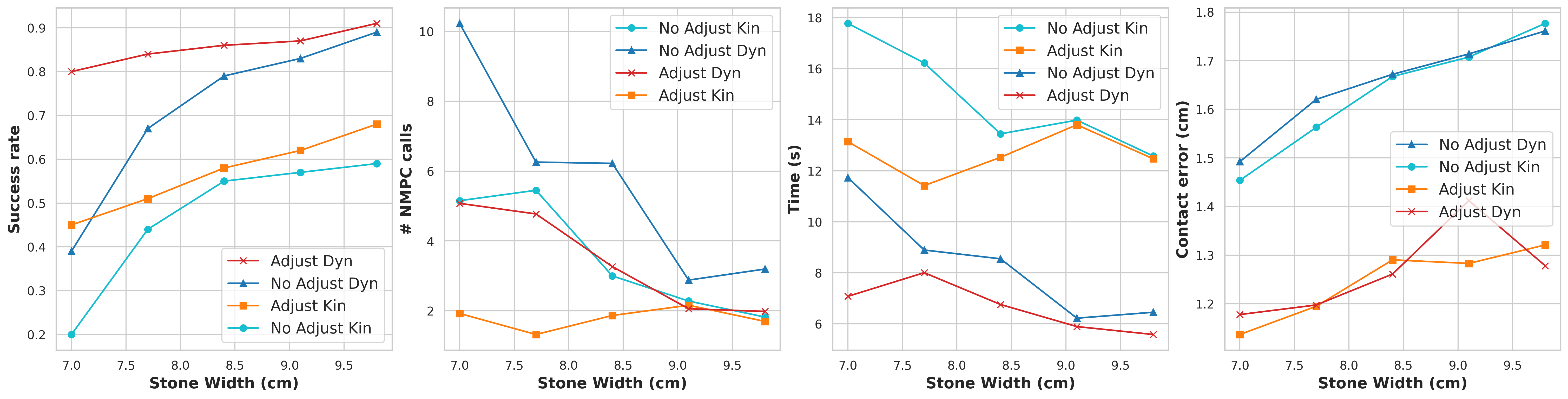}
    }
    \subfigure[Trotting Gait]{
        \includegraphics[width=0.97\textwidth]{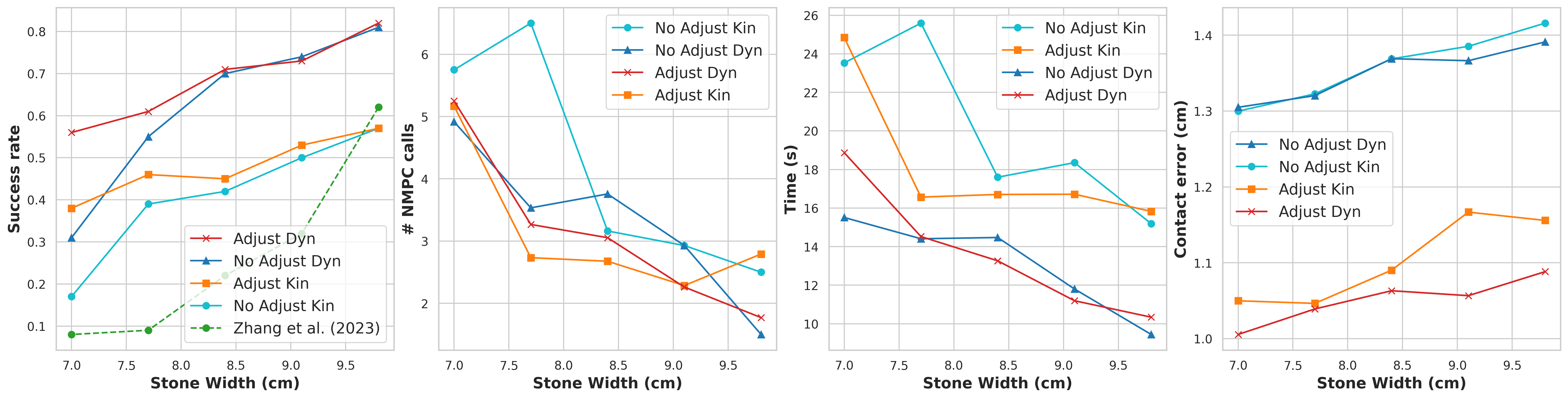}
    }
    \vspace{-2mm}
    \caption{Performance of the search for two different gaits. \textit{adjust}: experiments with target adjustment. \textit{kin}: experiments with kinematic pruning only. \textit{dyn}: experiments with kinematic and dynamic pruning.
    Note that we use the baseline heuristic for those experiments ($\alpha=0$, $\beta=0$). Results of~\cite{zhang2023learning} are added for comparison in trotting gate.} 
    \label{fig:main-graph}
    \vspace{-.8cm}
\end{figure}

We ran MCTS for a maximum of $10000$ iterations and stop the search when the first dynamically feasible contact plan is found (referred to as \textit{success} later).
We consider different metrics to evaluate the performance: the success rate, the number of NMPC simulations required to find a dynamically feasible solution (noted $\#\text{NMPC calls}$), search time, and the position error between achieved and targeted contact locations (noted $\text{contact error}$). We consider in the experiments randomized environment with different stepping stone sizes (side length ranging from $7 \text{ cm}$ to $9.7 \text{ cm}$). Statistics are averaged over $100$ randomly selected environments for each stepping stone size. Results are shown in Fig. \ref{fig:main-graph}. In the following we discuss the results.

\paragraph{MCTS Contact planner with target adjustment.}
\label{sec:results:adjustment}

For both gaits, adjusting the target contact locations improves the success rate. This improvement is more significant for stepping stones with a small radius. The success rate almost doubles on the smallest stones (with dynamic pruning). With the target adjustment network, the low-level controller is about $0.3 \text{ cm}$ more accurate on average on reaching the targeted contact locations.



\paragraph{MCTS Contact Planner with Dynamic Pruning.}
\label{sec:results:dymamic}

Dynamic pruning results in a more efficient search process.
As can be seen in Fig. \ref{fig:main-graph}, it reduces the number of NMPC simulations to perform compared to kinematic pruning only (on average $4.1$ against $5.8$ for jumping). It reduces the time of the search by more than 50\%, as time-consuming NMPC simulations are avoided and fewer nodes are added to the tree.
Although kinematic checks are fast, they often consider many transitions feasible that are not dynamically possible. Therefore, a significant portion of search time is wasted on expanding and selecting nodes that do not yield dynamically viable solutions. Indeed, we observed that the number of MCTS iterations is about $10$ times lower with dynamic pruning. Combined with target adjustment, dynamic pruning achieves the best success rate for both jumping and trotting gait.

\subsection{Comparison with state-of-the-art RL approach}
\label{sec:results:RL}

We replicate the work of \cite{zhang2023learning} to train a quadruped robot to traverse a stepping stones environment using reinforcement learning. Since the original code is not publicly available, we validated our implementation by reproducing comparable results with Anymal-D quadruped in Isaac Lab \cite{mittal2023orbit}.
We then adapted the environment and learning procedure to the Unitree Go2 robot. The training follows a two-stage process: first, the agent is pre-trained in a “Stones Everywhere” setup, followed by fine-tuning to configurations similar to our experimental setup described in \ref{sec:results:setup}. We reduced  the minimum stepping stone size to $12 \text{ cm}$. The agent was not able to learn for smaller stepping stones following the method of \cite{zhang2023learning}.
As shown in the Fig.~\ref{fig:main-graph}, the policy is not able to traverse stepping stones smaller than the one seen in training. On the smallest stones, our method have a success rate about $5$ times higher than the RL policy from \cite{zhang2023learning}.

\subsection{MCTS Contact planner with safety and accuracy heuristics}
\label{sec:results:safety}

We performed a grid search to find $\alpha$ and $\beta$ hyperparameters values that would improve the search performance (according to the metrics introduced before). Results on environments with the smallest stepping stone size are summarized in the Table \ref{tab:performance}. Chosen $(\alpha, \beta)$ values improve the search efficiency while keeping a similar success rate. This can be explained as the pruning strategy remains the same, and therefore, the tree search explores similar paths but in a different order. With the new heuristic, for the jumping gait, the number of NMPC calls is halved and the search time is divided by 3. Improvements were less significant for the trotting gait.

\begin{table}[h]
\vspace{-3mm}
\centering
\caption{Performance comparison with and without the proposed heuristic.}
\begin{tabular}{@{}lccccc@{}}
\toprule
\textbf{Algorithm} & \textbf{Success Rate} & \textbf{\#NMPC Calls} & \textbf{Time (s)} & \textbf{Iterations} & \textbf{Contacts Error (cm)} \\ \midrule
\multicolumn{6}{l}{\textbf{Jumping}} \\ \midrule
$\alpha=\beta=0$       & 80\%             & 5.27            & 12.05         & 270.4               & 1.25                \\
$\alpha=0.2,\beta=0.4$ & 77\%             & 2.54            & 4.61          & 107.7               & 1.14                \\ \midrule
\multicolumn{6}{l}{\textbf{Trotting}} \\ \midrule
$\alpha=\beta=0$       & 56\%             & 3.6             & 14.5          & 360.5               & 1.19                \\
$\alpha=0.4,\beta=0.2$ & 52\%             & 3.11            & 13.9          & 513.4               & 1.16                \\ \bottomrule
\end{tabular}
\vspace{-8mm}
\label{tab:performance}
\end{table}

\section{Conclusions and future works}\label{sec:conclusions}

This paper proposes a method to improve contact planning for legged robots in sparse environments. By combining model-based and learning-based approaches, the method efficiently explores the search space for feasible contact plans. Offline training is used to prune dynamically infeasible transitions and correct low-level controller inaccuracies at runtime.

Experiments on randomized stepping stones showed faster search times and higher success rates than the baseline. The framework’s precision enabled traversal of very small stepping stones, outperforming RL-based and MPC-based methods.

Future work includes implementing the framework on real robots, planning gait sequences for better performance, and extending it to loco-manipulation problems.


\bibliography{l4dc2025_latex_template/master}

\begin{thebibliography}{23}
\providecommand{\natexlab}[1]{#1}
\providecommand{\url}[1]{\texttt{#1}}
\expandafter\ifx\csname urlstyle\endcsname\relax
  \providecommand{\doi}[1]{doi: #1}\else
  \providecommand{\doi}{doi: \begingroup \urlstyle{rm}\Url}\fi

\bibitem[Aceituno-Cabezas et~al.(2017)Aceituno-Cabezas, Mastalli, Dai, Focchi, Radulescu, Caldwell, Cappelletto, Grieco, Fern{\'a}ndez-L{\'o}pez, and Semini]{aceituno2017simultaneous}
Bernardo Aceituno-Cabezas, Carlos Mastalli, Hongkai Dai, Michele Focchi, Andreea Radulescu, Darwin~G Caldwell, Jos{\'e} Cappelletto, Juan~C Grieco, Gerardo Fern{\'a}ndez-L{\'o}pez, and Claudio Semini.
\newblock Simultaneous contact, gait, and motion planning for robust multilegged locomotion via mixed-integer convex optimization.
\newblock \emph{IEEE Robotics and Automation Letters}, 3\penalty0 (3):\penalty0 2531--2538, 2017.

\bibitem[Amatucci et~al.(2022)Amatucci, Kim, Hwangbo, and Park]{amatucci2022monte}
Lorznzo Amatucci, Joon-Ha Kim, Jemin Hwangbo, and Hae-Won Park.
\newblock Monte carlo tree search gait planner for non-gaited legged system control.
\newblock In \emph{2022 International Conference on Robotics and Automation (ICRA)}, pages 4701--4707. IEEE, 2022.

\bibitem[Bogdanovic et~al.(2022)Bogdanovic, Khadiv, and Righetti]{bogdanovic2022model}
Miroslav Bogdanovic, Majid Khadiv, and Ludo Righetti.
\newblock Model-free reinforcement learning for robust locomotion using demonstrations from trajectory optimization.
\newblock \emph{Frontiers in Robotics and AI}, 9, 2022.

\bibitem[Bratta et~al.(2024)Bratta, Meduri, Focchi, Righetti, and Semini]{Bratta2024}
Angelo Bratta, Avadesh Meduri, Michele Focchi, Ludovic Righetti, and Claudio Semini.
\newblock Contactnet: Online multi-contact planning for acyclic legged robot locomotion.
\newblock In \emph{2024 21st International Conference on Ubiquitous Robots (UR)}, pages 747--754, 2024.
\newblock \doi{10.1109/UR61395.2024.10597477}.

\bibitem[Deits and Tedrake(2014)]{deits2014footstep}
Robin Deits and Russ Tedrake.
\newblock Footstep planning on uneven terrain with mixed-integer convex optimization.
\newblock In \emph{2014 IEEE-RAS international conference on humanoid robots}, pages 279--286. IEEE, 2014.

\bibitem[Deits and Tedrake(2015)]{deits2015computing}
Robin Deits and Russ Tedrake.
\newblock Computing large convex regions of obstacle-free space through semidefinite programming.
\newblock In \emph{Algorithmic Foundations of Robotics XI: Selected Contributions of the Eleventh International Workshop on the Algorithmic Foundations of Robotics}, pages 109--124. Springer, 2015.

\bibitem[Dhédin et~al.(2024)Dhédin, Ravi, Jordana, Zhu, Meduri, Righetti, Schölkopf, and Khadiv]{dhédin2024diffusionbasedlearningcontactplans}
Victor Dhédin, Adithya Kumar~Chinnakkonda Ravi, Armand Jordana, Huaijiang Zhu, Avadesh Meduri, Ludovic Righetti, Bernhard Schölkopf, and Majid Khadiv.
\newblock Diffusion-based learning of contact plans for agile locomotion, 2024.
\newblock URL \url{https://arxiv.org/abs/2403.03639}.

\bibitem[Grandia et~al.(2023)Grandia, Jenelten, Yang, Farshidian, and Hutter]{grandia2023perceptive}
Ruben Grandia, Fabian Jenelten, Shaohui Yang, Farbod Farshidian, and Marco Hutter.
\newblock Perceptive locomotion through nonlinear model predictive control.
\newblock \emph{IEEE Transactions on Robotics}, 2023.

\bibitem[Jenelten et~al.(2024)Jenelten, He, Farshidian, and Hutter]{jenelten2024dtc}
Fabian Jenelten, Junzhe He, Farbod Farshidian, and Marco Hutter.
\newblock Dtc: Deep tracking control.
\newblock \emph{Science Robotics}, 9\penalty0 (86):\penalty0 eadh5401, 2024.

\bibitem[Kocsis and Szepesv{\'a}ri(2006)]{kocsis2006bandit}
Levente Kocsis and Csaba Szepesv{\'a}ri.
\newblock Bandit based monte-carlo planning.
\newblock In \emph{European conference on machine learning}, pages 282--293. Springer, 2006.

\bibitem[Lin et~al.(2020)Lin, Righetti, and Berenson]{lin2020robusthumanoidcontactplanning}
Yu-Chi Lin, Ludovic Righetti, and Dmitry Berenson.
\newblock Robust humanoid contact planning with learned zero- and one-step capturability prediction, 2020.
\newblock URL \url{https://arxiv.org/abs/1909.09233}.

\bibitem[Meduri et~al.(2021)Meduri, Khadiv, and Righetti]{meduri2021deepq}
Avadesh Meduri, Majid Khadiv, and Ludovic Righetti.
\newblock Deepq stepper: A framework for reactive dynamic walking on uneven terrain.
\newblock In \emph{2021 IEEE International Conference on Robotics and Automation (ICRA)}, pages 2099--2105. IEEE, 2021.

\bibitem[Meduri et~al.(2023)Meduri, Shah, Viereck, Khadiv, Havoutis, and Righetti]{meduri2023biconmp}
Avadesh Meduri, Paarth Shah, Julian Viereck, Majid Khadiv, Ioannis Havoutis, and Ludovic Righetti.
\newblock Biconmp: A nonlinear model predictive control framework for whole body motion planning.
\newblock \emph{IEEE Transactions on Robotics}, 2023.

\bibitem[Mittal et~al.(2023)Mittal, Yu, Yu, Liu, Rudin, Hoeller, Yuan, Singh, Guo, Mazhar, Mandlekar, Babich, State, Hutter, and Garg]{mittal2023orbit}
Mayank Mittal, Calvin Yu, Qinxi Yu, Jingzhou Liu, Nikita Rudin, David Hoeller, Jia~Lin Yuan, Ritvik Singh, Yunrong Guo, Hammad Mazhar, Ajay Mandlekar, Buck Babich, Gavriel State, Marco Hutter, and Animesh Garg.
\newblock Orbit: A unified simulation framework for interactive robot learning environments.
\newblock \emph{IEEE Robotics and Automation Letters}, 8\penalty0 (6):\penalty0 3740--3747, 2023.
\newblock \doi{10.1109/LRA.2023.3270034}.

\bibitem[Mordatch et~al.(2012)Mordatch, Todorov, and Popovi{\'c}]{mordatch2012discovery}
Igor Mordatch, Emanuel Todorov, and Zoran Popovi{\'c}.
\newblock Discovery of complex behaviors through contact-invariant optimization.
\newblock \emph{ACM Transactions on Graphics (ToG)}, 31\penalty0 (4):\penalty0 1--8, 2012.

\bibitem[Ponton et~al.(2021)Ponton, Khadiv, Meduri, and Righetti]{ponton2021efficient}
Brahayam Ponton, Majid Khadiv, Avadesh Meduri, and Ludovic Righetti.
\newblock Efficient multicontact pattern generation with sequential convex approximations of the centroidal dynamics.
\newblock \emph{IEEE Transactions on Robotics}, 37\penalty0 (5):\penalty0 1661--1679, 2021.

\bibitem[Posa et~al.(2014)Posa, Cantu, and Tedrake]{posa2014direct}
Michael Posa, Cecilia Cantu, and Russ Tedrake.
\newblock A direct method for trajectory optimization of rigid bodies through contact.
\newblock \emph{The International Journal of Robotics Research}, 33\penalty0 (1):\penalty0 69--81, 2014.

\bibitem[Tassa et~al.(2012)Tassa, Erez, and Todorov]{tassa2012synthesis}
Yuval Tassa, Tom Erez, and Emanuel Todorov.
\newblock Synthesis and stabilization of complex behaviors through online trajectory optimization.
\newblock In \emph{2012 IEEE/RSJ International Conference on Intelligent Robots and Systems}, pages 4906--4913. IEEE, 2012.

\bibitem[Todorov et~al.(2012)Todorov, Erez, and Tassa]{Todorov2012}
Emanuel Todorov, Tom Erez, and Yuval Tassa.
\newblock Mujoco: A physics engine for model-based control.
\newblock In \emph{2012 IEEE/RSJ International Conference on Intelligent Robots and Systems}, pages 5026--5033, 2012.
\newblock \doi{10.1109/IROS.2012.6386109}.

\bibitem[Tonneau et~al.(2020)Tonneau, Song, Fernbach, Mansard, Taix, and Del~Prete]{tonneau2020slim}
Steve Tonneau, Daeun Song, Pierre Fernbach, Nicolas Mansard, Michel Taix, and Andrea Del~Prete.
\newblock Sl1m: Sparse l1-norm minimization for contact planning on uneven terrain.
\newblock In \emph{2020 IEEE International Conference on Robotics and Automation (ICRA)}, pages 6604--6610. IEEE, 2020.

\bibitem[Winkler et~al.(2018)Winkler, Bellicoso, Hutter, and Buchli]{winkler2018gait}
Alexander~W Winkler, C~Dario Bellicoso, Marco Hutter, and Jonas Buchli.
\newblock Gait and trajectory optimization for legged systems through phase-based end-effector parameterization.
\newblock \emph{IEEE Robotics and Automation Letters}, 3\penalty0 (3):\penalty0 1560--1567, 2018.

\bibitem[Zhang et~al.(2023)Zhang, Rudin, Hoeller, and Hutter]{zhang2023learning}
Chong Zhang, Nikita Rudin, David Hoeller, and Marco Hutter.
\newblock Learning agile locomotion on risky terrains.
\newblock \emph{arXiv preprint arXiv:2311.10484}, 2023.

\bibitem[Zhu et~al.(2023)Zhu, Meduri, and Righetti]{zhu2023efficient}
Huaijiang Zhu, Avadesh Meduri, and Ludovic Righetti.
\newblock Efficient object manipulation planning with monte carlo tree search.
\newblock In \emph{2023 IEEE/RSJ international conference on intelligent robots and systems (IROS)}. IEEE, 2023.

\end{thebibliography}



\end{document}